\documentclass[conference]{IEEEtran}
\IEEEoverridecommandlockouts

\usepackage{cite}
\usepackage{amsmath,amssymb,amsfonts}
\usepackage{algorithmic}
\usepackage{graphicx}
\usepackage{textcomp}
\usepackage{xcolor}

\usepackage{graphicx}
\usepackage{listings}
\usepackage{tcolorbox}

\usepackage{caption}
\usepackage{pdflscape}
\usepackage{subcaption}

\usepackage{lscape} 

\usepackage{algorithm}

\usepackage{xcolor}
\definecolor{darkergreen}{rgb}{0,0.5,0}
\definecolor{Orange}{RGB}{255,165,0}
\definecolor{Green}{RGB}{127,255,0}

\usepackage{url} 
\usepackage{hyperref}

\def\BibTeX{{\rm B\kern-.05em{\sc i\kern-.025em b}\kern-.08em
    T\kern-.1667em\lower.7ex\hbox{E}\kern-.125emX}}

\begin{document}

\title{A Flexible Multi-Agent LLM-Human Framework for Fast Human Validated Tool Building\\
}

\author{
  \IEEEauthorblockN{Xavier Daull}
  \IEEEauthorblockA{%
    \textit{Toulon Univ, Aix Marseille Univ, CNRS, LIS}\\
    Toulon, France\\
    xavier.daull@lis-lab.fr
  }
\and  
\IEEEauthorblockN{Patrice Bellot}
  \IEEEauthorblockA{
    \textit{Aix Marseille Univ, CNRS, LIS}\\
    Marseille, France\\
    patrice.bellot@univ-amu.fr}
\and
  \IEEEauthorblockN{Emmanuel Bruno}
  \IEEEauthorblockA{
    \textit{Toulon Univ, Aix Marseille Univ, CNRS, LIS}\\
    Toulon, France\\
    emmanuel.bruno@lis-lab.fr} 
\and
  \IEEEauthorblockN{Vincent Martin}
  \IEEEauthorblockA{
    \textit{Naval Group}\\
    Toulon, France\\
    vincent.martin@naval-group.com}
\and
  \IEEEauthorblockN{Elisabeth Murisasco}
  \IEEEauthorblockA{
    \textit{Toulon Univ, Aix Marseille Univ, CNRS, LIS}\\
    Toulon, France\\
    elisabeth.murisasco@lis-lab.fr
  }
}

\maketitle

\begin{abstract}
We introduce CollabToolBuilder, a flexible multi-agent LLM framework with expert-in-the-loop (HITL) guidance that iteratively learns to create tools for a target goal, aligning with human intent and process, while minimizing time for task/domain adaptation effort and human feedback capture. The architecture generates and validates tools via four specialized agents (Coach, Coder, Critic, Capitalizer) using a reinforced dynamic prompt and systematic human feedback integration to reinforce each agent's role toward goals and constraints. This work is best viewed as a system-level integration and methodology combining multi-agent in-context learning, HITL controls, and reusable tool capitalization for complex iterative problems such as scientific document generation. We illustrate it with preliminary experiments (e.g., generating state-of-the-art research papers or patents given an abstract) and discuss its applicability to other iterative problem-solving.

\end{abstract}


\section{Introduction}

Self-learning multi-agent LLMs and tool-making frameworks~\cite{caiLargeLanguageModels2023} have demonstrated promising capabilities in structured domains such as 3D sandbox games~\cite{wangVoyagerOpenEndedEmbodied2023, zhuGhostMinecraftGenerally2023}, sequential skill acquisition~\cite{colasAugmentingAutotelicAgents2023}, and mathematical discovery~\cite{romera-paredesMathematicalDiscoveriesProgram2023}. However, tackling ambiguous or non-factual problems requires additional multistep cognitive processes~\cite{luAIScientistFully2024,schmidgallAgentLaboratoryUsing2025}. These include collaborative agents’ reasoning~\cite{schmidgallAgentLaboratoryUsing2025,luAIScientistFully2024}, Chain-of-Thought problem solving~\cite{chuSurveyChainThought2023}, compositional question handling~\cite{hartillTeachingSmallerLanguage2023}, action planning~\cite{yaoReActSynergizingReasoning2022}, and multi-agent coordination~\cite{chenCoMMCollaborativeMultiAgent2024}.

We propose a conceptual framework and practical system that fuses multi-agent LLMs with an on-demand human-in-the-loop (HITL) mechanism to iteratively develop tools, and refine agents' roles, tested here for generating complex scientific synthesis. This poses unique challenges that require extensive domain knowledge and references, structured reasoning processes, iterative refinements, and robust fact-based validation, some elements that usual text generation methods often fail to address. Our approach works in modes ranging from fully autonomous to human-guided. It leverages automatic metrics to assess document structure, content quality, and reference accuracy, while incorporating expert feedback to iteratively refine the output.

Our contributions include: (1) a method to extend contextual tool development to real-world non-factual challenges via an interactive human-LLM collaboration applied to scientific document generation, producing human-validated, editable and reusable hybrid tools (e.g., mixing LLM inferences into code logics) for similar problems - open sourced as CollabToolBuilder; (2) a dataset and evaluation framework for complex document generation - open sourced as DEA; (3) a library to replace an existing LLM with a human-steered LLM - open sourced as HumanLLM; (4) "preliminary" reinforced dynamic prompt and macro-micro feedback strategies integrating systematic rich AI macro feedback with detailed human micro feedback reinforcing the AI agent's role.



\begin{figure*}[htbp]
    \includegraphics[width=\textwidth]{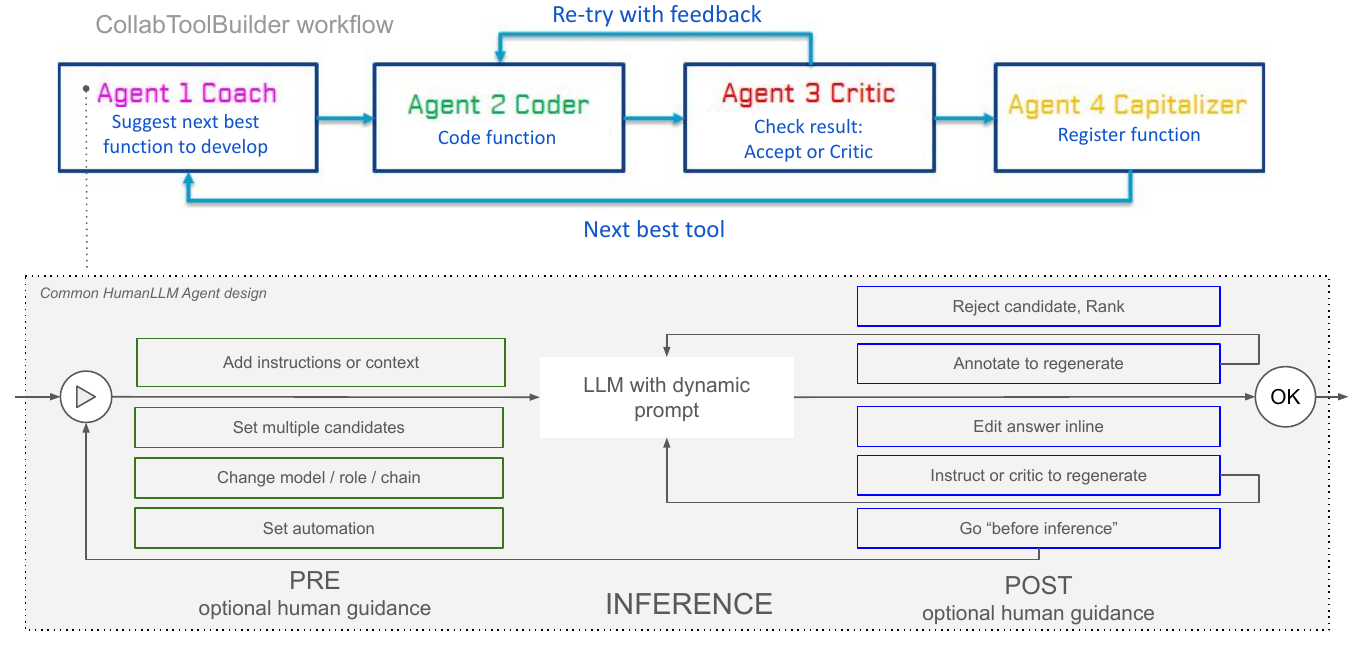}
    \caption{General tool building workflow and common internal workflow of each human guided LLM agent}
    \label{fig:HumanLLMAgents_Workflow}
\end{figure*}

\section{Related Work}

Early agentic systems such as LATM~\cite{caiLargeLanguageModels2023} and Voyager~\cite{wangVoyagerOpenEndedEmbodied2023} showed that a collaboration of role-specialized LLM agents can autonomously acquire tool libraries in simulated environments. Then ChatDev~\cite{qian2024chatdev} acting as a virtual software team (designer, coder, tester, reviewer) demonstrated end-to-end software development. Those approaches extended CAMEL’s role-playing agents~\cite{li2023camel} to cooperate through inception-prompted dialogue to achieve complex tasks, sequential skills (tools) acquisition~\cite{colasAugmentingAutotelicAgents2023}, mathematical discovery~\cite{romera-paredesMathematicalDiscoveriesProgram2023}, chain-of-thought reasoning~\cite{chuSurveyChainThought2023}, and multi-agent coordination~\cite{chenCoMMCollaborativeMultiAgent2024}. These studies confirmed that distributing subtasks among communicating LLM agents, often supported by external tools or helper code, improves scalability, factuality, and robustness over monolithic prompting.

An additional line of research focuses on “long-form content generation,” demonstrating how multiple LLMs can collaborate to improve coherence and factuality over single-pass prompting. For instance, CollabStory analyzes narrative consistency when up to 5 LLM “authors” jointly write fiction~\cite{venkatraman2024collabstory}, while Chain-of-Agents decomposes extended reasoning into communicating worker and manager agents~\cite{zhang2024chain}. WriteHere further scales this idea via hierarchical recursive planning to support arbitrarily long reports~\cite{yang2025docagent}. In the domain of technical writing, DocAgent employs a reader–writer–verifier topology to ensure accurate code documentation~\cite{yang2025docagent}. Storm combines multiple LLMs as dynamic, complementary expert perspectives to draft and organize Wikipedia-style articles, and CoStorm augments this setup with an optional human moderator to guide the LLM conversation. Although CoStorm introduces initial human-in-the-loop guidance, none of these systems produce reusable, human-validated tools that guarantee repeatable, human-aligned steps in document generation.

Benchmarks such as MultiAgentBench~\cite{zhu2025multiagentbenchevaluatingcollaborationcompetition} reveal that autonomous agents often drift when faced with ambiguous goals; conversely, hybrid systems like Agent Laboratory~\cite{schmidgallAgentLaboratoryUsing2025} or HULA~\cite{takerngsaksiriHULA2025} insert human expertise only at coarse checkpoints without feedback capitalization, missing opportunities for rapid iterative guidance.  Human reinforcement (RLHF) pipelines \cite{dongRLHFWorkflowReward2024,burnsWeakStrongGeneralizationEliciting2024,sunEasyHardGeneralizationScalable2024} and iterated-decomposition strategies \cite{reppertIteratedDecompositionImproving2023} provide complementary signals but stop short of producing structured, reusable documents.

\emph{We bridge these gaps with an architecture that :}
\emph{(i)} extends Voyager 4 agents architecture (Coach, Coder, Critic, Capitalizer)~\cite{wangVoyagerOpenEndedEmbodied2023} to document generation, and injects actionable human feedback at every agent loop both for immediate improvement and long-term agent role reinforcement,  
\emph{(ii)} equips the \textbf{Coach} with template-driven outline and planning, 
\emph{(iii)} extends the \textbf{Coder} to generate hybrid LLM\,+\,Python tools (see Figure \ref{fig:FunctionExample1}) for evidence gathering, and  
\emph{(iv)} empowers a domain-aware \textbf{Critic} that measures semantic coverage, citation accuracy, and plan structure compliance for scientific documents.  
This results in a transparent and human-aligned multi-agent workflow producing \emph{verifiable document generation tools}.

\section{Design of the Human LLM-Driven Tool building Framework}
\label{principles}

The architecture we propose showcases the collaborative dynamic between humans and LLM agents within a problem-solving environment, where four agents (Coach, Coder, Critic, Capitalizer) operate in a learning loop similar to Voyager~\cite{wangVoyagerOpenEndedEmbodied2023}, and humans can intervene as needed to supervise or refine the process, as illustrated in figure \ref{fig:HumanLLMAgents_Workflow}.
The LLM agents continuously identify and create the next best tool for the problem, given the current tool set. They also simultaneously self-refine their roles by enriching execution and feedback history into their dynamic prompt. The Coach identifies and specifies the best next tool, the Coder implements it and tests it on examples of problems, the Critic validates or instructs the Coder to improve, and the Capitalizer references successfully implemented tools as well as failed tools after retries (e.g., 'too hard'). This tool library informs the Coach of the learning evolution, providing a state of the current tool set. For learning and testing tools built, we simulate on a set of document examples the creation or evolution of the document content, or problem memory, toward the known solution when applying this tool. The problem's memory for each example begins with an empty solution and progresses toward a solved problem. It is used for evaluating the gain of a tool implementation, and for the Coach to identify the need for a "new" or "improved" or "specialized" tool given the current state of the different examples. This mechanism of iterative improvement of a tool library for a goal (e.g. scientific SOTA report creation) highly relies on initial human feedback to refine LLM agents' roles, under specified goals and constraints, unknown technical information, or specific knowledge. Human intervention at each LLM agent step, before or after inference and processing, can be a human proactive decision or reactive (queried) based on automatic triggers. Some human feedback may be permanent or re-introduced for example for critic validations that can’t be delegated to LLM agents due to responsibility concerns or because they lack access to or can’t evaluate necessary information (e.g. handling sensitive data or performing physical tests).

\emph{Learning task.} The system initially learns from a set of existing scientific papers as the target solution and measures the deviation from the target by computing semantic distances between the target and generated paper's structure (embeddings), contents, and references, weighted by their lengths. It ensures that the generated paper converges to a structure, content, and references similar to the target.

\emph{Agent's reinforced dynamic prompts.}
Each agent’s prompt is dynamic; we employ a mechanism which could be named a Reinforced Dynamic Prompt (RDP) – we design each agent’s prompt to be dynamic and cumulative, reinforced by execution results and feedback history. The prompt template is:
\begin{quote}
\{ROLE-GOAL-CONSTRAINTS\}\;
+\;\{STATE OBSERVATION\}\;
+\;\{TASK\}\;
+\;\{EXAMPLES\}
+\;\{FEEDBACKS\}\
\end{quote}
This prompt is updated after every iteration with new context. In our RDP schema, macro-level signals (e.g., automatic semantic scores or other quantitative feedback) are injected into the prompt’s state observation, while micro-level feedback (from human experts or an LLM self-critique) is appended to the feedback field with different strategies (e.g. diversity, age, negative, and positive...). In effect, each agent’s prompt becomes richer on every cycle, reinforcing the context and corrections from prior steps and closing the feedback loop between the AI agents’ outputs and expert guidance. The RDP's configuration of each agent can be manually set and/or optimized via Bayesian or generative optimization.
In accordance with the core feedback typology outlined by~\cite{metz2024mapping}, our framework already captures quantitative, comparative, corrective, and demonstrative feedbacks (4 out of the 5 core feedbacks), allowing us to construct a mix of few-shot prompt contexts through similarity, mutual information, and scoring strategies to be incorporated in a RDP.

\emph{Architecture design.} Generating long-form scientific documents (e.g. state-of-the-art surveys, patents, Wikipedia articles), resolving ambiguities, requires extensive bibliographic research, iterative reasoning, and precise formatting, tasks that challenge even state-of-the-art LLMs. In a departure from existing approaches, our original contribution proposes an iterative framework that uses human feedback to swiftly refine agents' roles to any field and goal, enabling efficient tool development with minimal search space and human input. Key components include:

\textbf{Problem Environment and Memory:} A simulation module instantiates each problem with its goal and available actions (OpenAI Gym format\footnote{https://github.com/openai/gym}). In the case of scientific documents and in order to help validate each tool built, a set of solved problems is given with a structured plan, contents, and resource lists. Semantic encodings of plans and content enable automatic feedback (via semantic distances weighted by length delta and priorities) and serve as a structured memory for testing tool-generated modifications.

\textbf{Macro-Micro Feedback Loop:} An automatic quantitative feedback (length and semantic similarity metrics on different aspects of content, structure and resources) is combined with qualitative evaluations (rule and LLM based evaluation on tool's output vs specifications) to iteratively refine outputs using this macro level of quantitative and qualitative evaluation.
Additionally (see bottom of figure \ref{fig:HumanLLMAgents_Workflow}), human agents can intervene "pre-inference" to adjust the agent's configuration or answer directly, and "post-inference" to adjust the agent's output by correcting, choosing among propositions, annotating or instructing corrections.
The LLM agents then derive both automatic macro feedback and optional fine-grained human feedback (micro) into immediate output correction on one side, and capitalize it in its pool of feedback for reinforced dynamic prompt on the other side.

\textbf{Tool Library Development:} Validated tools are stored in a dynamic semantic library~\cite{wangVoyagerOpenEndedEmbodied2023, zhuGhostMinecraftGenerally2023,colasAugmentingAutotelicAgents2023}. Each tool is enriched with performance metrics to facilitate efficient reuse, similar to \cite{yangLargeLanguageModels2023}.

\textbf{Inference, Planning, and Knowledge Integration:} The system decomposes complex requests into tasks, orchestrates tool usage, and executes code accordingly. A knowledge database —managed by neuro-symbolic code of tools— supports flexible knowledge mapping and access.

This streamlined architecture provides a scalable, feedback-driven framework for iterative tool development and complex problem-solving, effectively merging LLM capabilities with human guidance.

\section{Human Agents collaboration optimization}
\label{agent_section}
Making this tool building architecture general-purpose implies fast adaptation to problem, domain and constraints. Therefore, the loop of 4 LLM agents iteratively learns to develop by trial-and-error feedback (e.g. alignment, technical constraints, domain knowledge, etc.), automatically and optionally driven by human agents (in pre-guidance: instructs, configures before inference; in post-guidance: validates, corrects, selects...). The reliability of each LLM agent on their expected tasks over the acceptability of delegating fully or partially each task (e.g. risk, responsibility) should be a hard minimal constraint to assess acceptable automation level, then we estimate automation gain. Human guidance should ideally be optimized to maximize the impact of each intervention within its limited availability. We can model each agent, by adapting notation from~\cite{talebiradMultiAgentCollaborationHarnessing2023a}, as
\(\mathcal{A}_k=(T_k, R_k, S_k, C_k, F_k, \Gamma_k),\; 
k\in\{\text{coach},\text{coder},\text{critic},\text{capitalizer}\}\), where $T$ is its type (automatic, partial or full human), $R$ its role, $S$ its state,
$C$ its capabilities, $F$ the exposed functions, and $\Gamma$ its objective. The four agents \(\mathcal{A}_k\) will then operate in a loop where two conditional human-in-the-loop hooks govern the automation mode $T_k$: \textbf{HumanPreGuidance(\(\mathcal{A}_k\), context)} before each agent’s turn, and \textbf{HumanPostGuidance(\(\mathcal{A}_k\), output)} after inference.

 \noindent We choose whether to invoke human guidance on agent \(k\) by solving:
\[
\max_{X_i\in\{0,1\}} Z \;=\; \sum_{i=1}^n S_i\,X_i
\quad\text{s.t.}\quad
\begin{cases}
   \sum_{i=1}^n t_i X_i \le T,\\
   M_k \ge  R_k,\\
   X_i \le I_{i,\mathrm{trigger}}
\end{cases}
\]

 where: \(S_i\) is the predicted benefit of guiding \(\mathcal{A}_k\); \(X_i\) is the binary decision of human intervention; \(n\) is the number of possible human interventions; \(t_i\) human time cost; \(T\) total available time; \(M_k\) agent’s task success probability; \(R_k\) risk threshold (minimum acceptable success probability); \(I_{i,\mathrm{trigger}}\) boolean "need-guidance" flag.

 \noindent We call for human guidance if the expected benefit $S_i$ outweighs its cost and if the agent’s reliability $M_k$ falls below the risk reliability threshold $R_k$, focusing expert effort where it matters most.
 
 In practice, this means our framework can be configured to only request human input when it is likely to yield a significant gain in performance relative to the time spent and aligns with risks and responsibility constraints. In our implementation, the trigger is manually set or with basic heuristics; the impact score $S_i$ for each type of human intervention can be estimated from history. Since the effectiveness of human input depends on a rich context understanding, a generative optimizer~\cite{chengTraceNextAutoDiff2024a} informed by a rich knowledge and set of contextual information would be an interesting direction to optimize this problem with no or minimal past runs to refine this decision policy.

\section{Experimentation}
\label{experimentation}
\subsection{Use case and dataset construction}
 In this paper, we experiment with the proposed approach, focusing on the generation of scientific synthesis from the analysis and mining of scientific articles. We built a dataset of 30 documents (i.e. 10 arXiv survey papers, 10 scientific articles from Wikipedia, 10 European patents). For each, we extracted titles, abstracts, plans, full content, and references, generating semantic embeddings using the open-source e5-base-v2 model and OpenAI-AdaV2 private API. The dataset and its generation pipeline are publicly available\footnote{\url{https://anonymous.4open.science/r/82aea3fcb2c7a732fcbe73bc8566d4b97d75759e}} to repeat, and also apply it to new documents. The task is to create tools able to generate documents similar to solutions' plans, content, and bibliography for a given subject.

\emph{Task.}  
Given a \textbf{title + abstract}, the system must generate a full survey closely aligned in structure, content, length, and citations to a hidden reference document.

\emph{The workflow} then proceeds in four steps: (1) \textbf{Coach} drafts a tool specification, (2) \textbf{Coder} implements and tests the code, (3) \textbf{Critic} evaluates results and requests new attempts to code if needed, and (4) \textbf{Capitalizer} archives the validated tool—each iteration enriched by human feedback and automatic scoring.

\subsection{System implementation}
We implemented the above architecture and the HITL mechanism described in Section~\ref{agent_section}. Each agent is driven by dynamic prompts, code logic, and selectable inference engines. To optimize costs, after some tests, GPT-4o was selected for the Coach and Coder agents due to its superior task understanding and code generation capabilities. GPT-3.5 was initially used for the Critic and Capitalizer roles to balance performance with cost; it showed similar performance on those tasks. Experiments can be run in batch mode (using Optuna for Bayesian optimization\footnote{\url{https://optuna.readthedocs.io/}}) or interactively via a Web interface (built with the Monaco Editor\footnote{\url{https://microsoft.github.io/monaco-editor/}}), while \href{https://opensearch.org}{OpenSearch}\footnote{\url{https://opensearch.org}} serves as the back-end to store embeddings, tools, and feedback.

Some key features of the implemented system:\begin{itemize}
     \item \textbf{Common HITL mechanism} - lower part of figure \ref{fig:HumanLLMAgents_Workflow}:  a class standardizes how human feedback is integrated for any agent called HumanLLM. For each agent, it specializes its configuration and memory, and the interactions with human agents are unified in a common web frontend,
     \item \textbf{HITL Mechanism PRE inference}: human agents can change or adjust LLM prompts by automatic recommendations, by instruction, or manually, add instructions or information to be submitted to the agent with the prompt, adjust automation level (e.g., number of iterations of one agent without human intervention), provide direct answers instead of the LLM inference, adjust the number of propositions (number of parallel inferences), change the LLM or chain to be used for inference, and inspect past answers and feedback,
     \item \textbf{HITL Mechanism POST inference}: human agents can reject some propositions, request regeneration based on instructions, provide multiple annotations to be used for regeneration, modify output inline, score the output, or restart the process before inference with the possibility to change any configuration,
     \item \textbf{Online viewer \& editor}: users can directly view answers, modify them via a built-in editor (Monaco), and use action buttons and annotation tools.
     \item \textbf{Answers side-by-side comparison}: each LLM agent can generate multiple candidate answers (we used 2–3 for Coach, 2–4 for Coder, and 1 each for Critic and Capitalizer). The UI displays these candidates side-by-side for the human to compare and decide. This feature, along with the editor and HITL menu, was crucial in reducing human cognitive load compared to our initial terminal-based interface,
     \item \textbf{Coder agent validation}: any code generated by the Coder undergoes post-processing. We perform a syntax check, then run the code against custom tests. If no tests are provided, the system attempts to auto-generate tests by identifying the main function of the code. When tests fail, the agent can also apply automatic fixes using an LLM and Web Search or information retrieval,
     \item \textbf{Persistent storage}: a unified persistence layer (e.g., OpenSearch or ChromaDB) that captures both vector embeddings and structured metadata—tools, human annotations, retrieved documents, and experimental results—and exposes a semantic‐search interface so agents can seamlessly fetch contextually relevant knowledge to enrich their prompts.
     \item \textbf{Optimizer}: a Bayesian hyperparameter‐tuning class and some dedicated scripts, which adjust system‐wide and per‐agent parameters to maximize target objectives (semantic similarity, qualitative metrics, throughput) to accelerate convergence when tackling new problems.
 \end{itemize}

\subsection{Experiments settings}
\begin{landscape}
\begin{figure}[p]
  \vspace*{-0.5cm}              
  \centering
  \makebox[\linewidth]{%
    \begin{tcolorbox}[
      width=\linewidth,
      colframe=white, colback=white,
      boxrule=0.1pt, sharp corners,
      left=1mm, right=1mm,
      top=1mm, bottom=1mm      
    ]
      \begin{minipage}[t]{0.4\linewidth}
        \includegraphics[width=\linewidth]{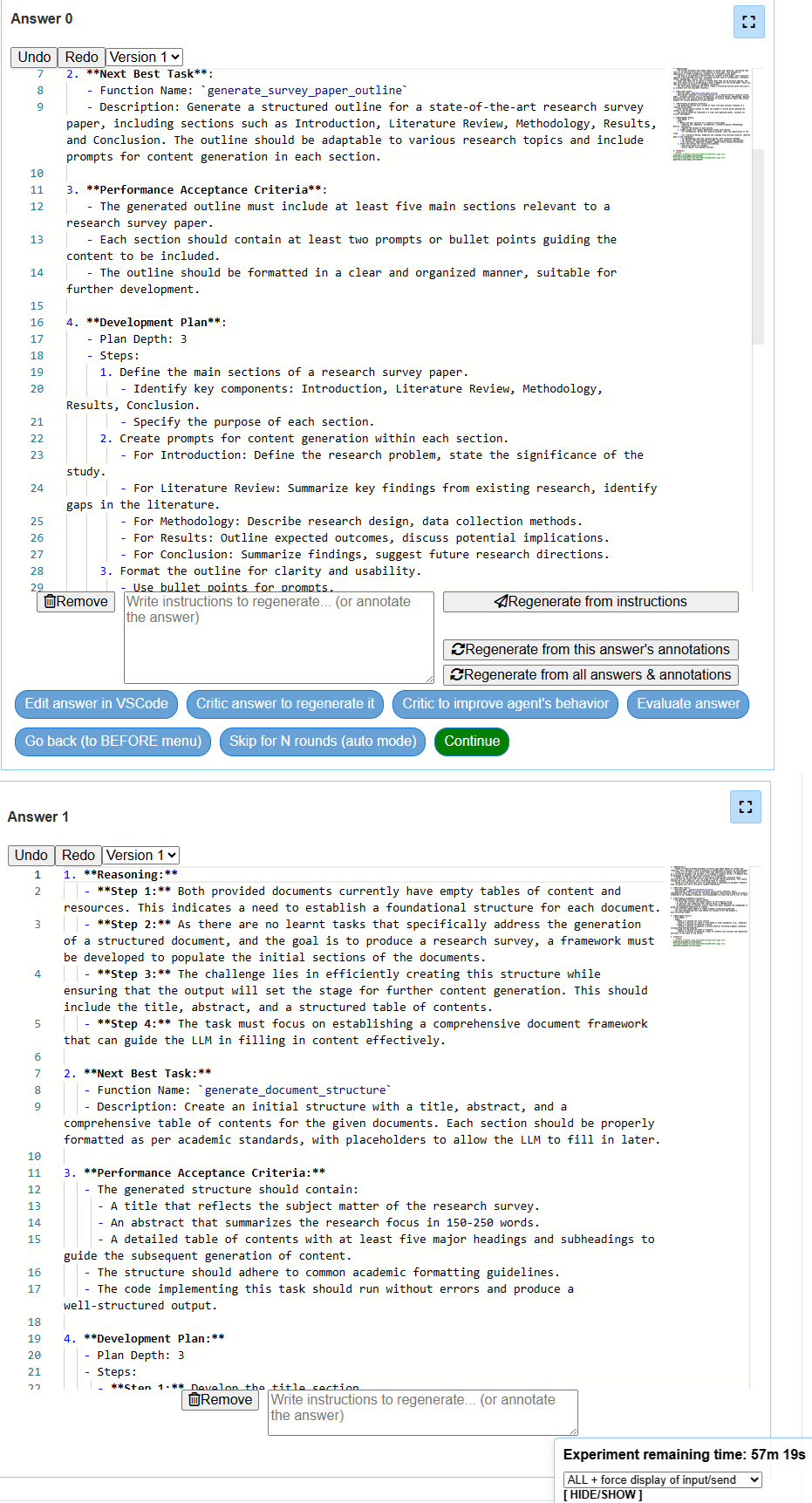}\\[4pt]
        \captionof{figure}{Coach step interface (1st iteration).}
        \label{fig:UI_Coach}
        \vspace{0.5em}\par\small
        The Coach agent converts 'title + abstract' into 2 tool specs. Human compares, edits, and annotates candidates to guide downstream agents.
      \end{minipage}\hfill
      \begin{minipage}[t]{0.45\linewidth}
        \includegraphics[width=\linewidth]{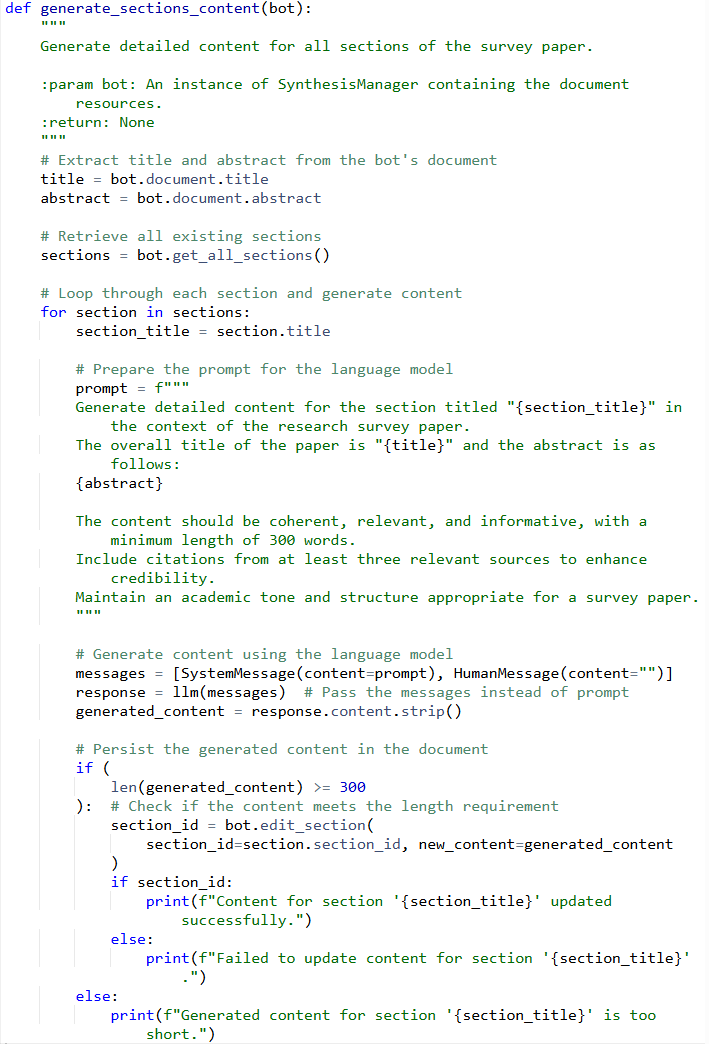}\\[4pt]
        \captionof{figure}{Capitalized function example (2nd iteration).}
        \label{fig:FunctionExample1}
        \vspace{0.5em}\par\small
        The function \texttt{generate\_sections\_content} was created and validated, including guards, test plan, and usage as future guidance.
      \end{minipage}
    \end{tcolorbox}%
  }
  \vspace*{-0.5cm}              
\end{figure}
\end{landscape}

\emph{Human experiments settings.}
Our human experiments involved 10 participants recruited from a team of young data scientists, with various levels of technical expertise (only one participant had expertise in LLM).
All human experiments were conducted following ethical guidelines. In 60-minute sessions (Figure \ref{fig:UI_Coach}), participants guided the system to develop “tools” (code in Python) iteratively transforming virtually empty documents into structured research papers. The cycle – comprising task specification, coding, validation, and capitalization – included both pre- and post-inference human interventions to correct errors and refine outputs. We measured the effectiveness of each tool by the semantic and structural similarity of its output to the target documents (e.g. existing scientific papers of 10 to 50 pages with proper plans, content, citations, and bibliography).

\emph{Automatic experiments settings.}
Our fully automatic configuration (OursAuto) employed Optuna over 200 trials to optimize key parameters, including prompt design, agents' configuration, temperature settings, and the maximum number of auto-fix attempts. Experiments comparing full HITL, full automation, and hybrid HITL-then-auto modes (see Table~\ref{tab:performances}) show that a partial human intervention strategy achieves the highest scores.

\emph{Evaluation metrics and baselines.}
Performance is measured using a composite score combining cosine similarity (plans, titles, content, references), coverage ratios, text length differences (see Table~\ref{tab:performances}, "Score Average"). The "Top Score" metric highlights the best-performing trials, while "Generated Codes" represents the total number of iterations before convergence. The global default score for Critic and optimizer, which can be redefined by humans depending on goals, is:
\[
\begin{array}{rl}
\mathrm{Score} = &
  w_{\mathrm{plan}}\,\widetilde{\mathrm{Sim}}_{\mathrm{plan}}
+ w_{\mathrm{title}}\,\widetilde{\mathrm{Sim}}_{\mathrm{titles}} \\
&+ w_{\mathrm{content}}\,\widetilde{\mathrm{Sim}}_{\mathrm{contents}}
+ w_{\mathrm{refs}}\,\widetilde{\mathrm{Sim}}_{\mathrm{refs}} \\
&+ w_{\mathrm{len}}\,\widetilde{\mathrm{Ratio}}_{\mathrm{contents.len}}
+ w_{\mathrm{cov}}\,\mathrm{Coverage}, \\
&\sum_{k} w_{k} = 1
\end{array}
\]

\section{Results \& Discussion}

\begin{table*}
\centering
\begin{tabular}{|l|c|c|c|c|c|c|}
\hline
\textbf{Run Type} & \textbf{Score Average} & \textbf{Top Score} & \textbf{Generated codes} & \textbf{Plan Similarity} & \textbf{Content Similarity} \\ \hline
60mn OursAuto & 36.8 & 38.8 & ~360 & 0.412 & 0.368 \\ \hline
60mn OursHITL & 33.5 & 34.1 & - & 0.368 & 0.412 \\ \hline
50mn OursHITL / 10mn OursAuto & \textbf{46.9} & \textbf{59.2} & ~60 & \textbf{0.567} & \textbf{0.674} \\ \hline
30mn OursHITL / 30mn OursAuto & 42.3 & 54.9 & ~180 & 0.487 & 0.496 \\ \hline
10mn OursAuto & 29.8 & 33.5 & ~60 & - & - \\ \hline
30mn OursAuto & 31.7 & 39.6 & ~180 & 0.391 & 0.444 \\ \hline
\end{tabular}
\caption{Scores over 3 runs, number of generated code candidates, similarities of plan and content to target documents on different time allocations and run types.}
\label{tab:performances}
\vspace{-5mm} 
\end{table*}

\begin{table}
\centering
\begin{tabular}{|l|p{2cm}|p{2cm}|c|}
\hline
\textbf{Agent} & \textbf{Human time allocation} & \textbf{LLM/Human time ratio} & \textbf{LLM choice} \\ \hline
Coach & 15\% & 1st:~50 next:~85\% & GPT-4o \\ \hline
Coder & 80\% & 1st:~40 next:~95\% & GPT-4o \\ \hline
Critic & 5\% & 1st/next:~90\% & GPT-3.5 \\ \hline
Capitalizer & \textasciitilde0\% & 1st/next:~100\% & GPT-3.5 \\ \hline
\end{tabular}
\caption{LLM and human involvement during learning phase - \emph{Human time allocation} is  the proportion of time spent at each agent step, summing to 100\% for the loop; \emph{LLM/Human time ratio} is how much LLM time is spent by LLM over human time at 1st iteration (Coach + Coder + Critic; then it might go back to Coder for a 2nd iteration to improve code, or to Capitalizer, then address a new task) and then for next iterations; \emph{LLM choice} is the type of LLM chosen by experience by humans for a good balance speed/performance/cost.}

\label{tab:human-contribution}
\end{table}

To understand the impact of human guidance, we evaluated our framework (Section~\ref{experimentation}) by comparing three configurations: (1) fully automated (OursAuto), (2) fully human-in-the-loop (OursHITL), and (3) a hybrid mode in which initial human guidance is followed by automation (OursHITL, then OursAuto). Table~\ref{tab:performances} summarizes the results in three runs for each configuration, reporting aggregated learning scores (based on semantic distances between generated and target documents), top scores, counts of generated code candidates, and similarity measures for document plans and content. Notably, the hybrid configuration achieved the highest performance (average score 46.9, top score 59.2; plan similarity 0.567 and content similarity 0.674 - references/citations score was not yet available at the time of this experiment but is now available in the released code), demonstrating that early human intervention helps steer the system toward better convergence.


Table~\ref{tab:human-contribution} provides a breakdown of human versus LLM involvement across the learning phases. This breakdown reveals that the majority of human effort was concentrated in the Coach (15\%) and Coder (80\%) stages. These are the stages of task definition and iterative code refinement, which aligns with the intuition that they are more complex. In these stages, the use of a more capable model (e.g., GPT-4o) significantly improved task alignment and code quality, while GPT-3.5 was sufficient for validation (critical) and archiving (Capitalizer). Meanwhile, validation (Critic) and archiving (Capitalizer) required minimal human time (5\% and 0\%). These findings highlight the complementary strengths of human expertise and LLM agents.

Beyond quantitative gains, we observed three qualitative effects that clearly emerged with humans in the loop.

\textbf{(1) Environment grounding.}
Human participants could instantly provide contextual knowledge that is
difficult (such as numerous trials and errors using unavailable libraries to uncover available ones, misuse of internal class methods) or impossible for an autonomous agent to infer, such as the type of accessible databases and visualization back-ends, or the location of API keys and credentials.  For instance, after one participant supplied the missing information, the Coder agent re-generated valid requests in a single iteration.

\textbf{(2) Intent sharpening.}
Humans refined underspecified or vague goals into precise functional
specifications, expected depth of research and generation, or unadapted approaches.

\textbf{(3) Fine-grained evaluation.}
{Critic} using automatic metrics and LLM analysis sometimes accepted syntactically valid and semantically misleading outputs, few human feedback quickly reinforces the expected output and uncovers some nuances. It corroborates findings from other HITL studies that emphasize the value of domain-aware feedback~\cite{takerngsaksiriHULA2025,schmidgallAgentLaboratoryUsing2025}. The agent quickly self-refines from few human feedback.

These observations confirm that human insight complements LLM
exploration—not merely by supplying rewards, as in RLHF~
\cite{dongRLHFWorkflowReward2024}, but by \emph{grounding} the process in a real technical ecosystem, refining intents and practices, and can inject missing knowledge into agents.

\emph{System warm-up:} we used Optuna (a Bayesian optimizer often used for hyperparameter tuning) for process stabilization over 200 trials of several key parameters - e.g. using a progressive temperature range (from 0 to 1.3) for the Coder agent balance solution exploration and determinism; limiting the number of automatic code fix attempts to 3 provided an effective trade-off between error correction and computational cost and time; restricting the available libraries (e.g., Langchain, Numpy, Pandas, Sklearn) reduces many initial errors at execution in the learning phase. Table \ref{tab:optuna-results} presents some key elements of the default configuration.

Initial runs revealed that seeding the process with code primitives (e.g. basic code examples demonstrating working code in the environment) significantly improved early performance, reduced search space, errors, and improved convergence toward target documents. Also, erroneous or misaligned human feedback at the early stage could propagate errors as a snowball effect, indicating a need for mechanisms to cancel or purge incorrect actions from system memory.

Overall, the experiments demonstrate that a hybrid human-LLM collaboration — especially one that leverages human guidance during initial iterations of agents Coach (guide on goal) and Coder (guide on environment constraints) — substantially improves tool generation and document synthesis time to stabilization and quality compared to either fully automated or fully HITL modes. These results validate the proposed methodology for iterative tool development (see tests in other domains in discussion) and suggest promising avenues for further refinement in error detection, prompt stabilization, and multi-agent coordination.

\begin{table*}
\centering
\begin{tabular}{|l|c|c|c|}
\hline
\textbf{Parameter} & \textbf{Optimal Value} & \textbf{Performance Metric} & \textbf{Notes} \\ \hline
Default common temperature & 0.5 & Highest code success rate & Excluded from future tuning \\ \hline
Progressive temperature range & 0 to 1.3 & Task score & Balance efficiency at 0 with different levels of creativity \\ \hline
Parallel code inferences & 3-4 & Gain in task score & Find most cost efficient setup ($>$4: gain not significant) \\ \hline
\texttt{max\_autofix} & 3 & Gain in task score & No significant gain beyond 3 \\ \hline
Libraries restriction & Langchain, Numpy, & Task score & Excluded complex libraries \\
 & Pandas, Sklearn &  &  \\ \hline
\end{tabular}
\caption{Extract of key system parameters tuned using Optuna over 200 trials}
\label{tab:optuna-results}
\end{table*}
    
\textbf{Discussion -}
\label{subsec:discussion}
Although the results are promising, they also highlight limitations. For instance, generating complete papers of equivalent quality to human-written would require to create or refine more tools with costly iterations, we had to limit tests. Also, generating tools via LLM agents chained in a multi-agent collaboration graph provides compositional power but increases complexity in coordination and end-to-end optimization. We need further experimentation to refine human interaction and end-to-end agents code optimization methods to improve the efficiency of this multi-agent graph. Using generative optimization~\cite{chengTraceNextAutoDiff2024a} with optional human-in-the-loop, rather than Bayesian optimization (Optuna), could allow the refinement of the agents code rather than just the agents parameters, with fewer iterations. Moreover, our approach's adaptability across diverse problem domains, such as generating scientific articles for the European patents, requires dataset quality improvement.

We also successfully tested this architecture on software issues solving (SWE-bench\footnote{\url{https://www.swebench.com}}), and marine pollution mitigation in a contest. It indicates that the approach also generalizes well even when an automatic cost or reward function for the target is not available by a domain expert feedback. For de-pollution, we configured the Coach agent to align with the chosen subject’s primary objective --marine de-pollution-- by incorporating their official mission statement. The Coder agent was provided with the list of data sources allowed in the contest, credentials to the database and visualization tools, and a chatbot for deploying generated functions. Lacking pre-solved cases, we set the goal distance metric to rely on systematic human feedback. Following initial tests with code errors related to missing API codes and URLs, the system successfully generated automatic weather updates and pollutant trajectory forecasts, displayed intervention locations, and plan recommendations. Domain experts confirmed the usefulness of these tools, and this first test confirmed the adaptability of this system to a very different domain and problem type even without an automatic cost or reward function to estimate the distance to the goal.

\section{Conclusion \& Future Work}

We have presented CollabToolBuilder, a human-in-the-loop multi-agent LLM framework that accelerates the creation of reusable, tool-based workflows for complex document synthesis. Preliminary experiments demonstrate the benefits of early expert guidance combined with automated optimization across scientific and environmental domains. The environment in our architecture is designed to be adaptable to various problem types (tested on scientific synthesis, marine pollution mitigation, software issues) by using automatic semantic distance or human evaluation. 

The dataset we release\footnote{\url{https://anonymous.4open.science/r/document_embedding_analysis-C581}} (research articles, patents, etc.) can guide the learning process to follow a structured reasoning plan, match semantic content and length, and justify with proper references.

Initial experiments highlighted several challenges. LLM agent's adaptation to a new task is accelerated but can remain difficult (unknown environment constraints, use of recent libraries). Collaboration efficiency was affected by the cognitive load on human agents, who often faced fatigue during iterative or long content generation. This limited the experimental scope at this stage: we aimed to focus on tasks where human agents have inner motivation for those difficult tasks. Finally, scaling this collaboration to include multiple human roles introduces anticipated coordination issues.

We now plan to:
(1) refine the collaboration framework between humans and LLM agents to allow advanced strategies of collaboration for tool and task adaptation across diverse domains, including exploring other scheduling of human intervention.
(2) continue HITL experiments (focusing on expert-in-the-loop) to reduce human cognitive load and LLM hallucinations, and to improve communication efficiency between human and LLM agents. We will also explore collaborations involving multiple human roles and different collaboration styles;
(3) explore LLM-based optimization~\cite{chengTraceNextAutoDiff2024a} for reinforced dynamic prompts and memory management;
(4) incorporate a versatile Planner agent and conduct broader ablations (including agent-removal settings) and larger-scale/industrial validations using the released pipeline.
\bibliographystyle{IEEEtran}
\bibliography{humanllmsynthesis_LONG-IEEE}

\end{document}